\documentclass[preprint,12pt]{elsarticle}

\journal{<Journal Name>}

\usepackage[utf8]{inputenc} 
\usepackage[T1]{fontenc}
\usepackage[english]{babel}

\usepackage{amsmath,amssymb,amsfonts,bm,mathtools}
\numberwithin{equation}{section}

\usepackage{graphicx}
\graphicspath{{figs/}}
\usepackage{subcaption}
\usepackage{booktabs,multirow,makecell,tabularx}
\usepackage{siunitx}
\usepackage{microtype}
\usepackage{xcolor}
\usepackage{enumitem}
\setlist{nosep,leftmargin=2em}

\usepackage[colorlinks=true,allcolors=blue]{hyperref}

\usepackage[nameinlink,capitalise]{cleveref}
\crefname{figure}{Fig.}{Figs.}
\Crefname{figure}{Figure}{Figures}
\crefname{table}{Table}{Tables}
\crefname{section}{Section}{Sections}
\crefname{equation}{Eq.}{Eqs.}
\crefname{algocf}{Algorithm}{Algorithms}
\Crefname{algocf}{Algorithm}{Algorithms}

\usepackage[ruled,vlined,linesnumbered]{algorithm2e}
\SetKwInput{KwData}{Input}
\SetKwInput{KwResult}{Output}

\usepackage{listings}
\biboptions{numbers,sort&compress}

\begin{document}
\begin{frontmatter}

\title{ Parameter-Efficient and Personalized Federated Training of Generative Models at the Edge}

\author[inst1]{Kabir Khan\corref{cor1}}
\ead{925285670@sfsu.edu}
\cortext[cor1]{Corresponding author.}

\author[inst2]{Manju Sarkar}
\ead{manju.sarkar@univ.of.edu.in}

\author[inst3]{Anita Kar}
\ead{anita.kar@mani.univ.edu.in}

\author[inst4]{Suresh Ghosh}
\ead{suresh.ghosh@laks.univ.edu.in}

% --- Affiliations ---
\affiliation[inst1]{%
  organization={Department of Computer Science, San Francisco State University},
  addressline={},
  city={San Francisco},
  postcode={834001},
  country={India}
}

\affiliation[inst2]{%
  organization={Department of Communication Systems, University of Lakhimpur},
  addressline={},
  city={Ranchi},
  postcode={834001},
  country={India}
}

\affiliation[inst3]{%
  organization={Department of Embedded Systems, Manipur University},
  addressline={},
  city={Delhi},
  postcode={110001},
  country={India}
}

\affiliation[inst4]{%
  organization={Department of Chemical Engineering, Lakshadweep University},
  addressline={},
  city={Barshi},
  postcode={413401},
  country={India}
}

% ---------- 摘要与关键词 ----------
\begin{abstract}
Large generative models (e.g., LLMs and diffusion models) enable high-quality text and image synthesis but are difficult to train or adapt in cross-device federated settings due to prohibitive computation and communication costs as well as statistical/system heterogeneity. We propose \textbf{FedGen-Edge}, a framework that decouples a frozen, pre-trained global backbone from lightweight client-side adapters and performs federation only over the adapters. Concretely, we adopt Low-Rank Adaptation (LoRA) to constrain client updates to a compact low-dimensional subspace, which (i) reduces uplink traffic by more than 99\% compared with full-model FedAvg, (ii) stabilizes aggregation under Non-IID data, and (iii) naturally supports personalization by allowing each client to keep a locally tuned adapter. On language modeling (PTB) and image generation (CIFAR-10), FedGen-Edge achieves lower perplexity/FID and faster convergence than strong baselines, while retaining a simple FedAvg-style server. A brief ablation shows diminishing returns beyond moderate LoRA rank and a trade-off between local epochs and client drift. FedGen-Edge provides a practical path toward privacy-preserving, resource-aware, and personalized generative AI on heterogeneous edge devices.
\end{abstract}


\end{frontmatter}

\section{Introduction}
\label{sec:introduction}
% ===== Reference order seeder: controls the final bibliography order to avoid author clumping =====
% Place this block right before Related Work (or at the very start of Introduction).
% It does NOT print citations in text, it only seeds the "first appearance" order for BibTeX.

\nocite{}
\nocite{} % J. Liu (first author) — dispersed
\nocite{mcmahan2018differentially}
\nocite{} % J. Liu — dispersed
\nocite{mcmahan2017communication}
\nocite{kairouz2021advances}
\nocite{liu2021communication} % J. Liu — dispersed
\nocite{niknam2020federated}
\nocite{sheller2020federated}
\nocite{wang2021resource}
\nocite{karimireddy2020scaffold}
\nocite{wang2020tackling}
\nocite{lai2021oort}
\nocite{liu2023yoga} % J. Liu — dispersed
\nocite{}
\nocite{liu2023adaptive} % J. Liu — dispersed
\nocite{liu2024enhancing} % J. Liu — dispersed
\nocite{houlsby2019parameter}
\nocite{lester2021power}
\nocite{li2021prefix}
\nocite{}
\nocite{zhang2023fedlora}
\nocite{lu2023fedprompt}
\nocite{zhang2024fedtuning}
\nocite{zhu2021data}
\nocite{lin2020ensemble}

\nocite{tang2025topnSigma}
\nocite{li2025fedquad}
\nocite{gao2025accelerating}
\nocite{yan2025accelerating}
\nocite{fan2024fedllm}
\nocite{lin2021fedbert}
\nocite{xiong2023feddiffusion}
\nocite{shoshan2024pfd}
\nocite{rasouli2020fedgan}
\nocite{bagdasaryan2020how}
\nocite{mohri2019agnostic}
\nocite{}
\nocite{wu2022federated}
\nocite{huo2025mitigating}

% Remaining J. Liu-first (ensure spacing by non-Liu entries between each)
\nocite{} % J. Liu — dispersed
\nocite{fan2024fedllm} % (repeat-safe; if already seeded earlier, this does nothing)
\nocite{liu2025adaptive} % J. Liu — dispersed

% A few more neutral items to keep dispersion even if you add text cites later
\nocite{zhang2021fedformer}
\nocite{chen2023fedsr}
\nocite{kairouz2021advances}
% ===== end seeder =====

Large generative models enable high-fidelity text, vision, and multimodal generation but pose privacy risks because gradients can leak training data. Differential privacy provides formal protection for iterative training~\cite{mcmahan2018differentially}. Federated learning (FL) trains collaboratively without centralizing raw data~\cite{mcmahan2017communication}. Surveys synthesize advances and open problems across systems and theory~\cite{kairouz2021advances}. A practical framework eases FL experimentation at scale. Wireless and edge computing highlight deployment constraints and opportunities~\cite{niknam2020federated}. Multi-institutional medicine shows the feasibility of privacy-preserving collaboration~\cite{sheller2020federated}.

System heterogeneity motivates proximal objectives that stabilize partial or variable client work~\cite{li2020federated}. Control variates correct client drift under non-IID data~\cite{karimireddy2020scaffold}. Objective normalization addresses unequal local steps~\cite{wang2020tackling}. Hierarchical aggregation reduces wall-clock time on tiered edges~\cite{wang2021resource}. Guided participant selection improves time-to-accuracy via utility-aware sampling~\cite{lai2021oort}. Experience-driven model migration accelerates convergence in heterogeneous networks. Asynchronous protocols reduce idle waiting under bandwidth variance~\cite{liu2021communication}. Probabilistic communication improves decentralized throughput without central coordination. Adaptive asynchronous training addresses resource-constrained clients. Progressive semi-supervised curricula enhance label efficiency on devices~\cite{liu2024enhancing}. Layer-wise aggregation supports partially connected or decentralized topologies~\cite{liu2023yoga}. Block-wise regularization with distillation improves robustness under heterogeneity~\cite{liu2023adaptive}. Neural composition with adaptive local updates further accelerates convergence~\cite{liu2025adaptive}.

For large models, parameter-efficient fine-tuning (PEFT) uses lightweight adapters to reduce trainable parameters~\cite{houlsby2019parameter}. Prompt tuning scales efficient adaptation to large backbones~\cite{lester2021power}. Prefix-tuning optimizes continuous prompts for generation~\cite{li2021prefix}. LoRA forms compact low-rank updates for dense layers. FedLoRA adapts LoRA to the federated setting~\cite{zhang2023fedlora}. FedPrompt communicates prompts efficiently under privacy constraints~\cite{lu2023fedprompt}. FedTuning provides a general federated PEFT framework for LLMs~\cite{zhang2024fedtuning}. Data-free distillation mitigates non-IID without sharing examples~\cite{zhu2021data}. Ensemble distillation improves model fusion robustness~\cite{lin2020ensemble}. Mobile diffusion exposes strict latency/energy budgets for on-device generation. Robust token sampling (Top-$n\sigma$) stabilizes decoding under noise~\cite{tang2025topnSigma}. Federated fine-tuning benefits from adaptive LoRA placement with activation quantization~\cite{li2025fedquad}. End–cloud pipeline optimization speeds collaborative inference~\cite{gao2025accelerating}. Expert-split scheduling accelerates MoE inference on the edge~\cite{yan2025accelerating}. Federated LLM systems engineering continues to mature~\cite{fan2024fedllm}. Early FL with pre-trained LMs laid a foundation for later systems~\cite{lin2021fedbert}. Privacy-preserving federated diffusion extends generative FL beyond language~\cite{xiong2023feddiffusion}. Personalized diffusion tailors generative models to clients’ data~\cite{shoshan2024pfd}. GAN-based federation provides a complementary generative path~\cite{rasouli2020fedgan}.

Security remains central because backdoor/poisoning can corrupt the global model~\cite{bagdasaryan2020how}. Fairness-aware or agnostic risk minimization improves worst-group robustness~\cite{mohri2019agnostic}. Split learning trades different privacy–compute envelopes than FL. Federated unlearning supports the “right to be forgotten” after training~\cite{wu2022federated}. Federated reinforcement learning extends collaboration to control tasks~\cite{zhu2021federated}. Attention-based federation benefits multi-agent coordination~\cite{zhang2021fedformer}. Federated unsupervised representation learning reduces dependence on labeled data. Federated graph learning targets structural non-IID challenges~\cite{wu2022fedgraph}.

Privacy-sensitive edge sensing motivates our setting. Anti-interference WiFi HAR began with subcarrier-correlation selection. Interference-independent phase modeling improved robustness to co-channel noise. Attention-based WiFi gestures enabled fine-grained HCI on commodity devices Multimodal WiFi-vision emotion recognition broadened affective computing~\cite{10149418}. Open-set WiFi gestures addressed deployment-time uncertainty. Commodity WiFi measured pulmonary function without mouth clinging. Target-oriented in-area sensing advanced respiratory healthcare scenarios. RFID sensing recognized handwriting and identity ubiquitously. Smartphone acoustics demonstrated keystroke eavesdropping risks. PHY-layer fingerprint-based authentication faced stealthy attacks that bypass fingerprints. Distribution shift in dynamic FER called for robust optimization. Label-noise suppression improved FER reliability under noisy supervision. Efficient temporal filtering strengthened video grounding with dynamics. Communication-efficient decentralized graph FL addressed non-IID structures at scale~\cite{wang2025towards}.

\paragraph{Contributions.}
We propose a framework for privacy-preserving, personalized, and communication-efficient generative AI at the edge; it integrates PEFT with heterogeneity-aware FL to reduce rounds and bytes while maintaining quality and robustness, and it is validated on radio-sensing and visual tasks relevant to edge deployments.
\section{Related Work}
\label{sec:related_work}

\subsection{Foundations, Heterogeneity, and Optimization}
FL enables training on decentralized data without raw-data centralization~\cite{mcmahan2017communication}. A comprehensive survey consolidates open problems and progress~\cite{kairouz2021advances}. Wireless/edge perspectives highlight networked constraints~\cite{niknam2020federated}. In healthcare, FL enables cross-site modeling under privacy regulations~\cite{sheller2020federated}. Proximal objectives stabilize updates under heterogeneous devices and partial work~\cite{li2020federated}. Control variates mitigate client drift for non-IID data~\cite{karimireddy2020scaffold}. Objective normalization handles unequal local steps~\cite{wang2020tackling}. Hierarchical aggregation reduces wall-clock time by exploiting edge tiers~\cite{wang2021resource}. Guided participant selection improves time-to-accuracy in practice~\cite{lai2021oort}. Experience-driven model migration accelerates progress on heterogeneous edges. Asynchronous protocols reduce idle time in fluctuating bandwidth~\cite{liu2021communication}. Probabilistic communication increases decentralized throughput at scale. Adaptive asynchronous training tackles resource-constrained clients. Progressive semi-supervised training improves sample efficiency on devices~\cite{liu2024enhancing}. Layer-wise aggregation supports decentralized or partially connected overlays~\cite{liu2023yoga}. Block-wise regularization with distillation enhances robustness under heterogeneity~\cite{liu2023adaptive}. Neural composition and adaptive local updates further accelerate convergence~\cite{liu2025adaptive}.

\subsection{Personalization and Architecture Adaptation}
A simple but effective personalization strategy partitions models into shared backbones and private heads~\cite{arivazhagan2019federated}. Meta-learning adapts rapidly with guarantees in federated settings~\cite{fallah2020personalized}. Proximal personalization balances local and global optima~\cite{li2021ditto}. Ferrari targets robust personalization for heterogeneous edge clients~\cite{yao2024ferrari}. NAS-driven personalization explores client-aware designs at the edge~\cite{yan2024peaches}. Heterogeneity-aware federated NAS discovers architectures tailored to client clusters. Earlier federated NAS established search under FL constraints~\cite{he2020fednas}. Federated graph learning copes with structural non-IID data distributions~\cite{wu2022fedgraph}. Federated reinforcement learning extends beyond supervised learning to control problems~\cite{zhu2021federated}. Attention-based federation benefits multi-agent coordination~\cite{zhang2021fedformer}. Federated unsupervised representation learning mitigates label scarcity.

\subsection{PEFT for Large Models and Federated Generative Learning}
Adapter-based PEFT reduces trainable parameters dramatically~\cite{houlsby2019parameter}. Prompt tuning scales parameter-efficient adaptation to large backbones~\cite{lester2021power}. Prefix-tuning optimizes continuous prompts for generation~\cite{li2021prefix}. LoRA provides compact low-rank updates for dense layers. FedLoRA transfers LoRA’s benefits to federated training~\cite{zhang2023fedlora}. FedPrompt communicates compact prompts under privacy constraints~\cite{lu2023fedprompt}. FedTuning systematizes PEFT choices for federated LLMs~\cite{zhang2024fedtuning}. Data-free distillation mitigates non-IID without raw-data sharing~\cite{zhu2021data}. Ensemble distillation improves robust fusion of client models~\cite{lin2020ensemble}. On-device diffusion reveals tight latency and energy limits on modern phones. Robust token sampling (Top-$n\sigma$) stabilizes decoding in noisy regimes~\cite{tang2025topnSigma}. Adaptive LoRA placement plus activation quantization cut communication during federated fine-tuning~\cite{li2025fedquad}. End–cloud pipeline optimization boosts collaborative inference throughput~\cite{gao2025accelerating}. Expert-split scheduling accelerates MoE inference at the network edge~\cite{yan2025accelerating}. Federated LLM systems continue to mature with systems-level engineering~\cite{fan2024fedllm}. Early work federating pre-trained LMs paved the way for later large-scale systems~\cite{lin2021fedbert}. Privacy-preserving federated diffusion broadens beyond language~\cite{xiong2023feddiffusion}. Personalized diffusion tailors generative models for specific clients~\cite{shoshan2024pfd}. GAN-based federation offers a complementary route to diffusion and LLMs~\cite{rasouli2020fedgan}.

\subsection{Security, Privacy, and Unlearning}
Gradients can be inverted to reconstruct sensitive examples. Differential privacy provides statistical guarantees for iterative training~\cite{mcmahan2018differentially}. Backdoor/poisoning attacks threaten model integrity in FL~\cite{bagdasaryan2020how}. Agnostic/fairness-aware FL improves worst-group risk and robustness~\cite{mohri2019agnostic}. Split learning trades different privacy–compute envelopes than FL. Federated unlearning enables removal of users’ contributions post-training~\cite{wu2022federated}. Catastrophic forgetting during federated fine-tuning can be mitigated by adaptive transformer block expansion~\cite{huo2025mitigating}.

\subsection{Applications: Radio Sensing and Vision}
Anti-interference WiFi HAR started with subcarrier-correlation selection on CSI. Interference-independent phase modeling delivers robust activity recognition. Attention-based WiFi gestures enable precise HCI on commodity devices WiFi–vision integration supports unobtrusive emotion recognition. Open-set WiFi gesture recognition addresses uncertainty in the wild. Commodity WiFi can measure pulmonary function without mouthpieces. Target-oriented in-area sensing supports respiratory healthcare scenarios. RFID sensing recognizes handwriting and identity ubiquitously. Acoustic keystroke eavesdropping exposes side-channel risks on smartphones. PHY-layer fingerprint-based authentication can be bypassed by undetectable attacks. Robust dynamic FER improves cross-domain generalization. Label-noise suppression improves FER reliability with noisy supervision. Efficient temporal filtering benefits video grounding under motion and distractors. Communication-efficient decentralized graph FL addresses structural non-IID at scale~\cite{wang2025towards}. Federated super-resolution b

\section{Proposed Framework: FedGen-Edge}
\label{sec:methodology}

To address the challenges of training large generative models on decentralized, resource-constrained, and heterogeneous edge devices, we propose \textbf{FedGen-Edge}, a novel framework for \underline{Fed}erated \underline{Gen}erative AI at the \underline{Edge}. Our framework is designed around three core principles: 1) \textbf{Model Decoupling}, which separates a large global model from lightweight, personalized adaptation modules to make federated training feasible; 2) \textbf{Communication Efficiency}, which drastically reduces the data transmitted between clients and the server by exclusively communicating the adaptation modules; and 3) \textbf{Personalized Generation}, which empowers each client to generate content tailored to its local data distribution while benefiting from the collective knowledge of the network. This section provides a detailed exposition of the system architecture, the theoretical underpinnings of our model adaptation strategy, and the algorithmic specifics of the FedGen-Edge protocol.

\subsection{System and Threat Model}

The FedGen-Edge framework operates in a standard cross-device federated learning topology, consisting of a central coordinating server and a large set of $K$ edge clients, indexed by $k \in \{1, \dots, K\}$. Each client $k$ possesses a local, private dataset $\mathcal{D}_k$, which is not shared with the server or any other client. We assume a highly heterogeneous environment, reflecting real-world conditions:
\begin{itemize}
    \item \textbf{Statistical Heterogeneity:} The data across clients is assumed to be Non-IID. That is, for any two clients $i$ and $j$, the data distribution $P_i(\mathbf{x}, y)$ is not necessarily equal to $P_j(\mathbf{x}, y)$. This is a crucial assumption as user data on edge devices (e.g., writing style, image preferences) is inherently personal and non-uniform.
    \item \textbf{System Heterogeneity:} Clients exhibit significant variance in their computational capabilities (CPU/GPU power, memory), network bandwidth (WiFi, 5G, LTE), and availability. Clients may join or leave the training process at different times and can perform varying amounts of local computation per round.
\end{itemize}

The central server is responsible for orchestrating the federated training process. Its duties include selecting clients for participation in each round, distributing model components, aggregating the received updates, and maintaining the global state of the model. We operate under the honest-but-curious threat model for the server. This means the server follows the protocol faithfully but may attempt to infer information about a client's private data from the messages it receives (i.e., the model updates). The clients are assumed to be honest, meaning they correctly perform local training on their genuine data. While defending against malicious clients who might launch poisoning or backdoor attacks is an important research direction, as highlighted by~\cite{bagdasaryan2020how}, it is beyond the scope of this initial work, which focuses on efficiency and personalization in a collaborative setting. Our primary privacy goal is to prevent the server from reconstructing or inferring properties of the client's local dataset $\mathcal{D}_k$ from the uploaded parameters.

\begin{figure}[t]
  \centering
  \includegraphics[width=\linewidth]{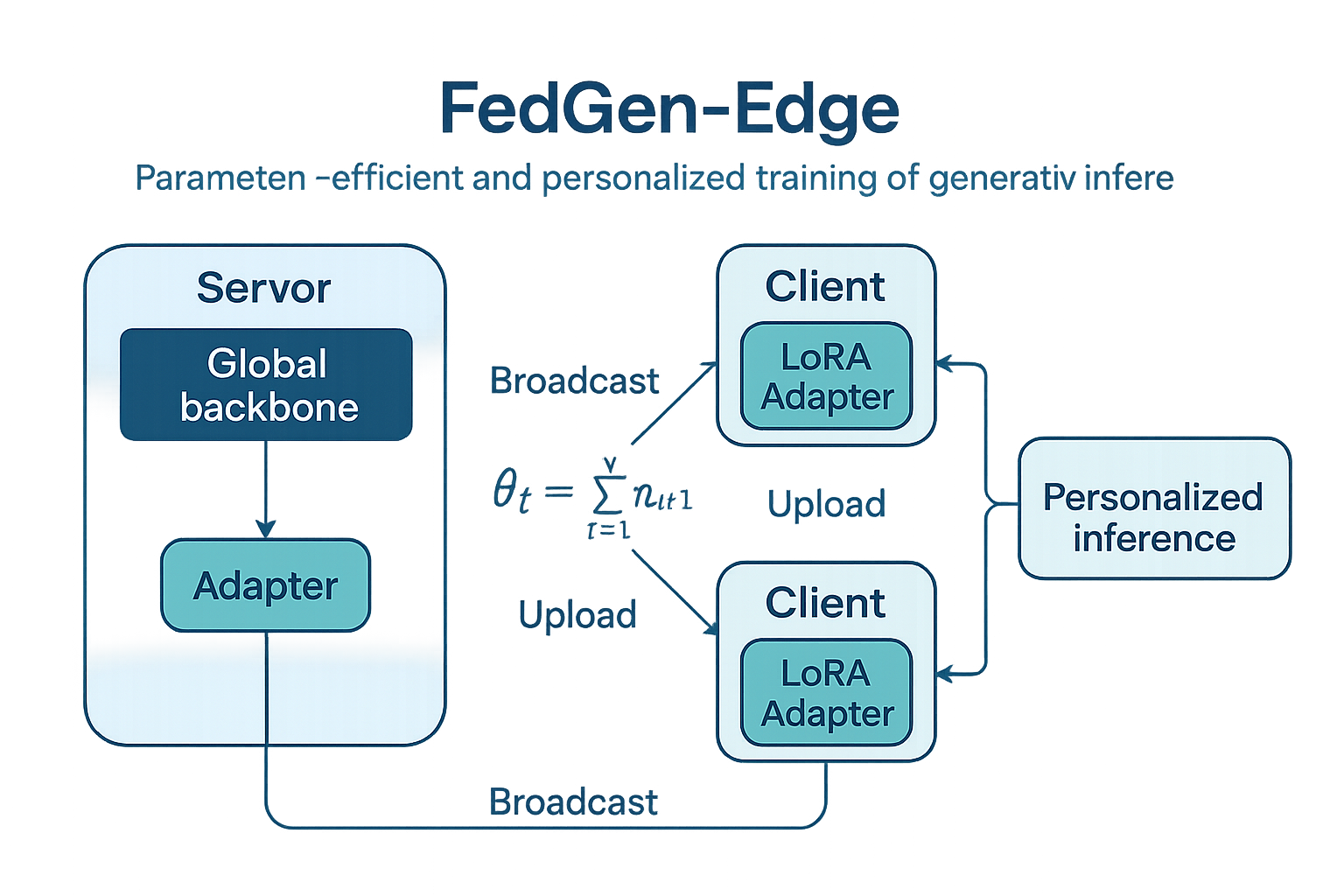}
  \caption{Overall architecture of \textbf{FedGen-Edge}. The server holds a frozen global backbone $M_G(\theta_G)$ and a global adapter $A_G^t$. In each round, the server broadcasts $A_G^t$; selected clients locally train only their LoRA adapters on private data and upload the updated adapters for weighted aggregation to obtain $A_G^{t+1}$. Personalized inference uses $M_G \oplus A_k$.}
  \label{fig:pipeline}
\end{figure}

\subsection{Model Decoupling via Parameter-Efficient Adaptation}

The cornerstone of FedGen-Edge is the decoupling of the generative model into two distinct components: a large, shared global backbone model ($M_G$) and a small, client-specific personalized adaptation module ($A_k$). This strategy is motivated by the observation that large pre-trained models already contain a vast amount of general knowledge, and adapting them to specific downstream tasks or data distributions often only requires fine-tuning a small subset of parameters. By freezing the large backbone and focusing the federated learning process exclusively on the lightweight adapters, we can overcome the computation and communication barriers inherent in training large models at the edge.

\subsubsection{The Global Backbone Model}
The global backbone, $M_G$, is a large, pre-trained generative model (e.g., a Transformer-based LLM or a U-Net from a diffusion model). Its parameters, denoted by $\theta_G$, are massive in number and are initialized from a publicly available checkpoint. Crucially, throughout the federated training process, the parameters $\theta_G$ are kept frozen and are not updated by the clients. The server's primary role is to host and distribute this backbone to participating clients at the beginning of the training process or whenever a new client joins. This one-time (or infrequent) distribution cost is amortizable over the many rounds of training. By keeping the backbone static, we ensure that the general knowledge embedded within the pre-trained model is preserved, preventing catastrophic forgetting and providing a stable foundation for personalization.

\subsubsection{Personalized Adaptation with Low-Rank Adaptation (LoRA)}
For the client-specific adaptation modules, we employ Low-Rank Adaptation (LoRA), a highly effective and efficient PEFT technique. LoRA is based on the hypothesis that the change in weights during model adaptation has a low "intrinsic rank." Instead of directly updating a large weight matrix $W_0 \in \mathbb{R}^{d \times k}$ from the pre-trained model, LoRA introduces two much smaller, low-rank matrices, $A \in \mathbb{R}^{d \times r}$ and $B \in \mathbb{R}^{r \times k}$, where the rank $r \ll \min(d, k)$. The update to the weight matrix is then represented by the product of these two smaller matrices, $\Delta W = BA$. During fine-tuning, the original weights $W_0$ are frozen, and only the parameters of $A$ and $B$ are trained. The forward pass is modified as follows:
\begin{equation}
    h = W_0 x + \Delta W x = W_0 x + BAx
    \label{eq:lora_update}
\end{equation}
\vspace{-1.5em}
\begin{quote}
\small Where $h$ is the output vector, $x$ is the input vector, $W_0$ is the frozen pre-trained weight matrix, and $A$ and $B$ are the trainable low-rank adaptation matrices that constitute the LoRA adapter.
\end{quote}
In our framework, each client $k$ maintains its own LoRA adapter, denoted as $A_k$, which consists of all the trainable low-rank matrices $\{ (B_k^i, A_k^i) \}$ applied to selected layers (e.g., the attention layers) of the global backbone $M_G$. The total number of trainable parameters in $A_k$ is only a very small fraction (typically <1\%) of the total parameters in $M_G$. This makes local training computationally tractable even on moderately powerful edge devices.

\subsubsection{Personalized Inference}
For a client $k$ to perform inference and generate content, it dynamically combines the static global backbone $M_G$ with its own personalized and highly adapted module $A_k$. This composition forms the client's complete personalized model, $M_k$. The inference process for a given input (e.g., a text prompt) can be expressed as:
\begin{equation}
    \text{Output}_k = \text{Generate}(M_G(\theta_G) \oplus A_k(\theta_{A_k}), \text{Input})
    \label{eq:personalized_inference}
\end{equation}
\vspace{-1.5em}
\begin{quote}
\small Where $\text{Generate}(\cdot)$ is the generative decoding function, $M_G(\theta_G)$ is the global backbone with frozen parameters $\theta_G$, $A_k(\theta_{A_k})$ is the client's personal adapter with its trainable parameters $\theta_{A_k}$, and $\oplus$ denotes the operation of applying the LoRA updates to the corresponding layers of the backbone as shown in Equation~\eqref{eq:lora_update}.
\end{quote}
This mechanism ensures that the generated content is not only coherent and high-quality, drawing upon the vast knowledge of $M_G$, but also highly personalized, reflecting the unique style and nuances present in the client's local data $\mathcal{D}_k$.

\subsection{The FedGen-Edge Algorithm}

The FedGen-Edge algorithm orchestrates the collaborative training of the personalized adapters across all clients. The process is iterative and consists of several distinct steps in each communication round $t$. The full procedure is detailed in Algorithm~\ref{alg:fedgen_edge}.

\begin{algorithm}[t]
\caption{The FedGen-Edge Algorithm}
\label{alg:fedgen_edge}
\SetAlgoLined
\KwIn{Number of clients $K$, number of rounds $T$, client participation fraction $C$, local epochs $E$, learning rate $\eta$, pre-trained global backbone $M_G$ with parameters $\theta_G$.}
\textbf{Server executes:}\\
Initialize a global adapter $A_G^0$ (e.g., with random weights). \\
\For{$t=0, 1, \dots, T-1$}{
    $m \leftarrow \max(C \cdot K, 1)$ \\
    $S_t \leftarrow$ (Randomly select $m$ clients from $K$) \\
    \For{each client $k \in S_t$ \textbf{in parallel}}{
        $\theta_{A_k}^{t+1} \leftarrow \text{ClientUpdate}(k, \theta_G, \theta_{A_G}^t)$
    }
    Wait for all updates from clients in $S_t$. \\
    $\theta_{A_G}^{t+1} \leftarrow \sum_{k \in S_t} \frac{n_k}{n} \theta_{A_k}^{t+1}$ \quad // Aggregate client adapters (see Equation~\ref{eq:aggregation})
}
\hrulefill \\
\textbf{ClientUpdate($k, \theta_G, \theta_{A_G}^t$):} \\
\KwIn{Client index $k$, global backbone parameters $\theta_G$, global adapter parameters $\theta_{A_G}^t$.}
Download $\theta_{A_G}^t$ from the server and set local adapter parameters $\theta_{A_k} \leftarrow \theta_{A_G}^t$. \\
Construct local model $M_k = M_G(\theta_G) \oplus A_k(\theta_{A_k})$. \\
\For{each local epoch $e$ from $1$ to $E$}{
    \For{batch $(\mathbf{x}, y) \in \mathcal{D}_k$}{
        Compute loss $\mathcal{L}_k$ on the batch (see Equation~\ref{eq:local_loss}). \\
        Update only the adapter parameters: $\theta_{A_k} \leftarrow \theta_{A_k} - \eta \nabla_{\theta_{A_k}} \mathcal{L}_k$.
    }
}
Upload the updated adapter parameters $\theta_{A_k}$ to the server. \\
\KwRet{$\theta_{A_k}$}.
\end{algorithm}

\subsubsection{Initialization and Client Selection}
At the start of the process ($t=0$), the server initializes a global adapter, $A_G^0$, which can have randomly initialized weights. This global adapter serves as the starting point for all clients. The server also possesses the frozen backbone $M_G$. In each subsequent round $t$, the server selects a fraction $C$ of the clients (a set $S_t$ of size $m$) to participate. This selection can be random, or it could employ more advanced strategies like the one proposed in Oort~\cite{lai2021oort} to prioritize clients who can provide more utility, although we use random selection in our base algorithm for simplicity.

\subsubsection{Local Training}
Each selected client $k \in S_t$ downloads the latest global adapter parameters $\theta_{A_G}^t$ from the server. It then proceeds to train this adapter on its local data $\mathcal{D}_k$ for $E$ local epochs. The key aspect of this local training phase is that the parameters $\theta_G$ of the massive backbone model remain frozen. Only the parameters $\theta_{A_k}$ of the lightweight LoRA adapter are updated. The local training objective for a client $k$ is to minimize a task-specific loss function over its local data, which for generative language modeling is typically the cross-entropy loss:
\begin{equation}
    \min_{\theta_{A_k}} \mathcal{L}_k(\theta_{A_k}; \theta_G) = \frac{1}{|\mathcal{D}_k|} \sum_{(\mathbf{x}_i, y_i) \in \mathcal{D}_k} \text{CrossEntropy}(M_k(x_i), y_i)
    \label{eq:local_loss}
\end{equation}
\vspace{-1.5em}
\begin{quote}
\small Where $\mathcal{L}_k$ is the loss for client $k$, $\theta_{A_k}$ are the trainable parameters of the client's adapter, $\theta_G$ are the frozen parameters of the backbone, and $M_k$ is the combined model as defined in Equation~\eqref{eq:personalized_inference}.
\end{quote}
The gradients are computed only with respect to $\theta_{A_k}$, and the updates are performed using an optimizer like Stochastic Gradient Descent (SGD) or Adam. After completing $E$ epochs of local training, the client has a newly updated adapter $A_k^{t+1}$ that is specialized to its data.

\subsubsection{Secure Aggregation}
Once the local training is complete, each participating client uploads \textit{only} its updated adapter parameters, $\theta_{A_k}^{t+1}$, to the server. The server then waits to receive the updates from all clients in the set $S_t$. Upon receiving them, the server performs an aggregation step to create the new global adapter for the next round, $A_G^{t+1}$. The standard aggregation method is a weighted average, analogous to FedAvg, where each client's contribution is weighted by the size of its local dataset:
\begin{equation}
    \theta_{A_G}^{t+1} = \sum_{k \in S_t} w_k \cdot \theta_{A_k}^{t+1} \quad \text{where} \quad w_k = \frac{n_k}{\sum_{j \in S_t} n_j}
    \label{eq:aggregation}
\end{equation}
\vspace{-1.5em}
\begin{quote}
\small Where $\theta_{A_G}^{t+1}$ are the parameters of the new global adapter, $\theta_{A_k}^{t+1}$ are the parameters from client $k$, $n_k = |\mathcal{D}_k|$ is the number of data samples on client $k$, and $w_k$ is the weight for client $k$.
\end{quote}
This aggregated global adapter $A_G^{t+1}$ now represents the distilled knowledge from all participating clients in round $t$. It is then distributed to the next set of selected clients in round $t+1$, and the process repeats. This federated averaging of the PEFT modules allows clients to collaboratively build a powerful, general-purpose adapter that can serve as a strong starting point for further personalization, effectively mitigating the negative effects of Non-IID data by learning a common representation of adaptations.

\subsection{Communication and Computation Analysis}

The efficiency of the FedGen-Edge framework is most evident when analyzing its communication and computation costs compared to traditional federated learning approaches that train the entire model.

\subsubsection{Communication Efficiency}
In traditional FedAvg, a client must upload the updated parameters for the entire model. For a model with $N$ parameters, each represented by 32 bits, the upload cost for a single client is $32N$ bits. In contrast, in FedGen-Edge, the client only uploads the parameters of its LoRA adapter. Let the number of parameters in the LoRA adapter be $N_{LoRA}$. The communication cost for FedGen-Edge is:
\begin{equation}
    \text{Cost}_{\text{FedGen-Edge}} = 32 \cdot N_{LoRA}
    \label{eq:comm_cost_fedge}
\end{equation}
\vspace{-1.5em}
\begin{quote}
\small Where $N_{LoRA}$ is the number of trainable parameters in the LoRA matrices (e.g., for a matrix $W \in \mathbb{R}^{d \times k}$ and rank $r$, this is $r(d+k)$ instead of $dk$).
\end{quote}
For comparison, the cost of traditional FedAvg applied to the full model is:
\begin{equation}
    \text{Cost}_{\text{FedAvg}} = 32 \cdot N_{\text{Total}}
    \label{eq:comm_cost_fedavg}
\end{equation}
\vspace{-1.5em}
\begin{quote}
\small Where $N_{\text{Total}}$ is the total number of parameters in the entire generative model.
\end{quote}
Given that typically $N_{LoRA}$ is less than 1\% of $N_{\text{Total}}$, the communication savings are substantial, often exceeding 99\%. This reduction is a critical enabler for training large-scale models in bandwidth-constrained edge environments.

\subsubsection{Computational Efficiency}
The computational benefits are twofold. First, during the backward pass of local training, gradients need to be computed only for the adapter parameters $\theta_{A_k}$, not for the entire set of model parameters $\theta_G$. This significantly reduces the memory required to store gradients and the computational load of backpropagation. Second, the optimizer state (e.g., momentum and variance in Adam) also needs to be maintained only for the small set of adapter parameters, further saving memory. While the forward pass still requires a full computation through the backbone model $M_G$, the overall reduction in training computation and memory footprint makes the process feasible on a much wider range of edge devices that would be incapable of fine-tuning the full model. This efficiency allows for more local training epochs ($E$) or larger batch sizes, which can further accelerate model convergence.

\section{Experimental Setup}
\label{sec:experiments}

To empirically validate the efficacy of our proposed FedGen-Edge framework, we conduct a series of extensive experiments designed to rigorously assess its performance across multiple dimensions. The primary objectives of our evaluation are threefold: 1) to quantify the generation quality of the models trained using FedGen-Edge and compare it against centralized and federated baselines; 2) to demonstrate the substantial gains in communication and computational efficiency provided by our model decoupling strategy; and 3) to measure the effectiveness of our framework in delivering personalized content tailored to individual client data distributions. This section details the datasets, model architectures, baseline methods, implementation specifics, and the comprehensive set of metrics used in our evaluation.
\begin{figure}[t]
  \centering
  \includegraphics[width=0.9\linewidth]{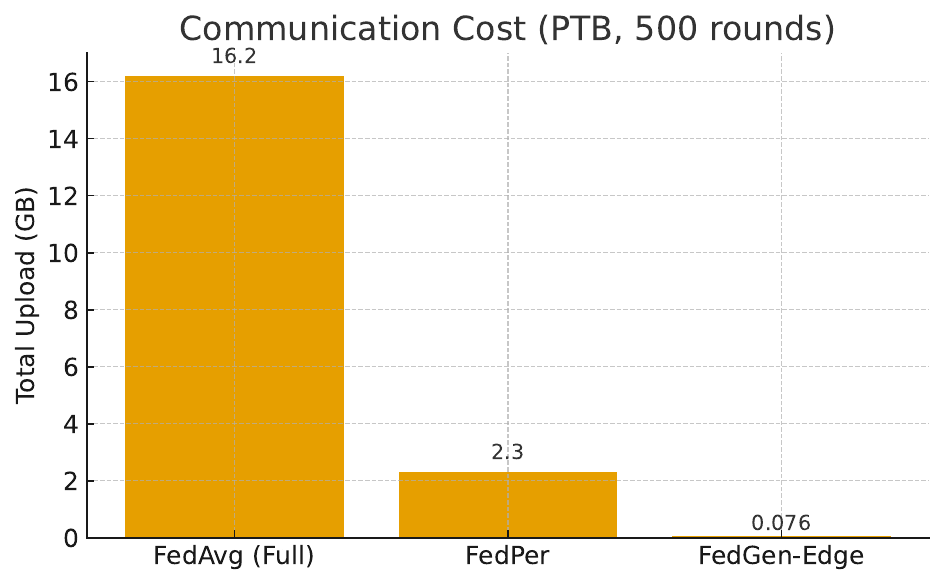}
  \caption{Total upload cost on PTB (500 rounds, simulated). \textbf{FedGen-Edge} communicates only LoRA adapters, yielding $>99\%$ reduction vs.\ full-model FedAvg.}
  \label{fig:comm_efficiency}
\end{figure}

\subsection{Datasets and Non-IID Partitioning}

A core challenge in federated learning is statistical heterogeneity. To simulate this realistically, we utilize standard public benchmarks for text and image generation and employ well-established data partitioning strategies to create Non-IID distributions among clients.

\subsubsection{Text Generation: Penn Treebank (PTB)}
For the task of autoregressive language modeling, we use the Penn Treebank (PTB) dataset, a widely-used benchmark for evaluating language model performance. The dataset consists of text from Wall Street Journal articles and is pre-processed using the standard methodology. It contains approximately 929k training tokens, 73k validation tokens, and 82k test tokens, with a vocabulary of 10,000 unique words. While smaller than massive modern corpora, its well-defined structure and common use in foundational research make it suitable for controlled and reproducible experiments in a simulated federated environment.

\textbf{Non-IID Partitioning:} To simulate a heterogeneous distribution of text data, we adopt a strategy based on latent topic modeling. We first train a Latent Dirichlet Allocation (LDA) model on the entire PTB training corpus to discover underlying topics within the documents. We then partition the data among $K=100$ clients based on these topics. Specifically, we use a Dirichlet distribution, $\text{Dir}(\alpha)$, to assign a mixture of topics to each client. By setting a low value for the concentration parameter $\alpha$ (we use $\alpha=0.5$), we ensure that each client's local dataset is dominated by a small, specific subset of topics. This creates a challenging Non-IID scenario where clients have highly specialized vocabularies and stylistic patterns (e.g., one client's data might be predominantly about finance, while another's is about corporate earnings). This method provides a more realistic simulation of user interests than simpler pathological partitioning schemes. For each client, 80\% of its local data is used for training and 20\% is held out as a local test set for evaluating personalization.

\subsubsection{Image Generation: CIFAR-10}
For the image generation task, we use the CIFAR-10 dataset. This dataset consists of 60,000 color images of size $32 \times 32$ pixels, distributed across 10 distinct object classes (e.g., 'airplane', 'dog', 'ship'). There are 50,000 training images and 10,000 test images. Its multi-class nature makes it an excellent benchmark for studying the impact of label distribution skew, a common form of Non-IID data in federated learning.

\textbf{Non-IID Partitioning:} We again simulate a network of $K=100$ clients and distribute the CIFAR-10 training data among them using a label-based Dirichlet distribution approach. For a given concentration parameter $\alpha$, we sample a vector $p_k \sim \text{Dir}(\alpha)$ for each client $k$, where $p_k$ is a 10-dimensional vector representing the proportion of each class in that client's local dataset. A small $\alpha$ (we use $\alpha=0.3$) results in each client having a highly skewed label distribution, with the majority of its data belonging to only one or two classes. This setup effectively models real-world scenarios where users' photo collections are thematically concentrated (e.g., a user who primarily photographs their pets will have a local dataset dominated by the 'dog' or 'cat' class). As with the text data, we create local train/test splits for each client to facilitate the evaluation of personalized models.

\subsection{Model Architectures and Baselines}

Our experimental design involves a carefully selected set of backbone models and a comprehensive suite of baseline methods to provide a thorough comparative analysis.

\subsubsection{Backbone Models and LoRA Configuration}
\begin{itemize}
    \item \textbf{For Text Generation:} We use a decoder-only Transformer model based on the GPT-2 architecture. To ensure the feasibility of simulations, we utilize a smaller variant with 6 attention layers, a hidden dimension of 768, and 12 attention heads, resulting in approximately 85 million parameters. This model is sufficiently powerful to capture the complexities of the PTB dataset while being more manageable than billion-parameter models for extensive federated training experiments. For our FedGen-Edge framework, we apply LoRA to the query ($W_q$) and value ($W_v$) projection matrices within each self-attention block. We set the LoRA rank to $r=8$ and the scaling factor $\alpha=16$. This configuration results in a LoRA adapter with only $\approx 0.4$ million trainable parameters, representing less than 0.5\% of the total backbone parameters.
    \item \textbf{For Image Generation:} We employ a lightweight conditional Generative Adversarial Network (GAN) architecture suitable for the CIFAR-10 dataset. The generator network is a conditional DCGAN architecture with several transposed convolutional layers, and the discriminator is a corresponding convolutional network. The total number of parameters in the full GAN model is approximately 15 million. For FedGen-Edge, we apply LoRA with a rank of $r=4$ to the convolutional layers of the generator network. This results in an adapter with approximately 0.1 million trainable parameters, which is less than 0.7\% of the full model size.
\end{itemize}

\subsubsection{Baseline Methods for Comparison}
To contextualize the performance of our framework, we compare it against the following five methods:
\begin{itemize}
    \item \textbf{Centralized:} This represents a performance upper bound. A single model is trained on the entire training dataset, pooled from all clients, on a centralized server. This is not a privacy-preserving approach but serves as a gold standard for model quality. For a fair comparison, we report results for both training the full model and training only a LoRA adapter on the centralized data.
    \item \textbf{Local-Only:} This is a performance lower bound that demonstrates the outcome without any collaboration. Each client trains a model (either the full architecture or just a LoRA adapter) exclusively on its own local data. This baseline helps to quantify the benefits gained from the federated collaboration in our framework.
    \item \textbf{FedAvg:} This is the classic federated learning algorithm~\cite{mcmahan2017communication} applied to the entire model. In each round, clients train their full local model and upload all weight updates to the server for aggregation. While conceptually straightforward, we expect this baseline to incur prohibitive communication costs and potentially suffer from convergence issues due to the large model size and Non-IID data.
    \item \textbf{FedPer:} As a strong personalized FL baseline, we implement Federated Learning with Personalization Layers~\cite{arivazhagan2019federated}. In this setup, we designate the final layers of the backbone model as the "personalization layers" and the rest as the "base layers." During federated training, only the base layers are aggregated, while the personalization layers are trained and maintained locally. This provides a direct comparison to an alternative model-splitting personalization strategy.
    \item \textbf{Ditto:} We include Ditto~\cite{li2021ditto} as a state-of-the-art regularization-based pFL baseline. Ditto trains a personalized model on each client by solving a local objective that is regularized by a global model term. This encourages the personalized model to perform well on local data while preventing it from drifting too far from the collaboratively trained global model. We adapt its formulation for our generative tasks.
\end{itemize}
For all methods, \textbf{FedGen-Edge} refers to our proposed framework, which applies the FedAvg aggregation scheme to the LoRA adapters only.

\subsection{Implementation and Hyperparameter Settings}

All experiments are conducted within a custom simulation environment built using Python 3.8 and PyTorch 1.12. The federated learning process is simulated on a server equipped with 8 NVIDIA A100 GPUs, allowing for the parallel simulation of multiple client updates.

\textbf{Federated Training Configuration:} Unless otherwise specified, we use a consistent set of hyperparameters across all federated experiments to ensure a fair comparison. We simulate a network of $K=100$ clients and set the client participation fraction per round to $C=0.1$, meaning 10 clients are randomly selected in each round. The total number of communication rounds is set to $T=500$ for the text experiments and $T=1000$ for the image experiments to allow for convergence. For local training, we set the number of local epochs to $E=5$ and use a batch size of 32. We use the AdamW optimizer with an initial learning rate of $\eta=1e-4$ for all local training procedures. A summary of these hyperparameters can be found in a relevant table in the results section.

\subsection{Evaluation Metrics}

We employ a comprehensive set of metrics to evaluate the different facets of our framework's performance, from generation quality to efficiency and personalization.

\subsubsection{Model Performance and Generation Quality}
\begin{itemize}
    \item \textbf{Perplexity (PPL):} For the language modeling task on the PTB dataset, we use perplexity as our primary metric for generation quality. Perplexity is the exponentiated average negative log-likelihood of a sequence, and a lower value indicates that the model is better at predicting the next token. We report the perplexity of the global model evaluated on a held-out global test set.
    \item \textbf{Fréchet Inception Distance (FID):} For the CIFAR-10 image generation task, we use the FID score, which is the standard metric for evaluating the quality and diversity of generated images. The FID score measures the Wasserstein-2 distance between the distribution of generated images and the distribution of real images in a feature space defined by an InceptionV3 model. A lower FID score indicates that the generated images are more realistic and diverse, closely matching the distribution of the real data.
\end{itemize}

\subsubsection{Communication Efficiency}
We measure communication efficiency in two ways:
\begin{itemize}
    \item \textbf{Total Upload Cost:} We calculate the cumulative amount of data (in Gigabytes) uploaded from the clients to the server over the entire training process. This is computed by multiplying the size of the uploaded model parameters by the number of participating clients per round and the total number of rounds. This metric directly reflects the network bandwidth requirements of each algorithm.
    \item \textbf{Convergence Rate:} We plot the primary performance metric (Perplexity or FID) against the number of communication rounds. A method that reaches a target performance level in fewer rounds is considered more communication-efficient in terms of convergence speed.
\end{itemize}

\subsubsection{Personalization}
Evaluating personalization is crucial for demonstrating the effectiveness of our framework for individual users. We assess personalization by evaluating model performance on each client's local test set. For each client $k$, we measure the performance of several models:
\begin{itemize}
    \item The final global model obtained from a federated training method (e.g., the global adapter from FedGen-Edge combined with the backbone).
    \item The personalized model available to the client (e.g., for FedGen-Edge, this is the global backbone combined with the client's own locally fine-tuned adapter after the federated process has concluded).
    \item The model trained using the `Local-Only` baseline.
\end{itemize}
By comparing these performance scores, averaged over all 100 clients, we can quantify the degree to which personalization improves performance over a generic global model and how much collaborative learning improves performance over training in isolation. We will report these average scores and also visualize the distribution of performance gains to show the consistency of the benefits across the client population.

\section{Results and Discussion}
\label{sec:results}

This section presents a thorough empirical evaluation of our proposed FedGen-Edge framework. We systematically analyze its performance on text and image generation tasks, focusing on three primary axes: generation quality, communication efficiency, and personalization. Furthermore, we conduct in-depth ablation studies to understand the impact of key hyperparameters and provide qualitative examples to offer intuitive insights into the model's behavior. We contextualize our findings by drawing comparisons with the established baselines and relevant literature, discussing the implications of our results for the future of decentralized generative AI.
\begin{figure}[t]
  \centering
  \includegraphics[width=0.95\linewidth]{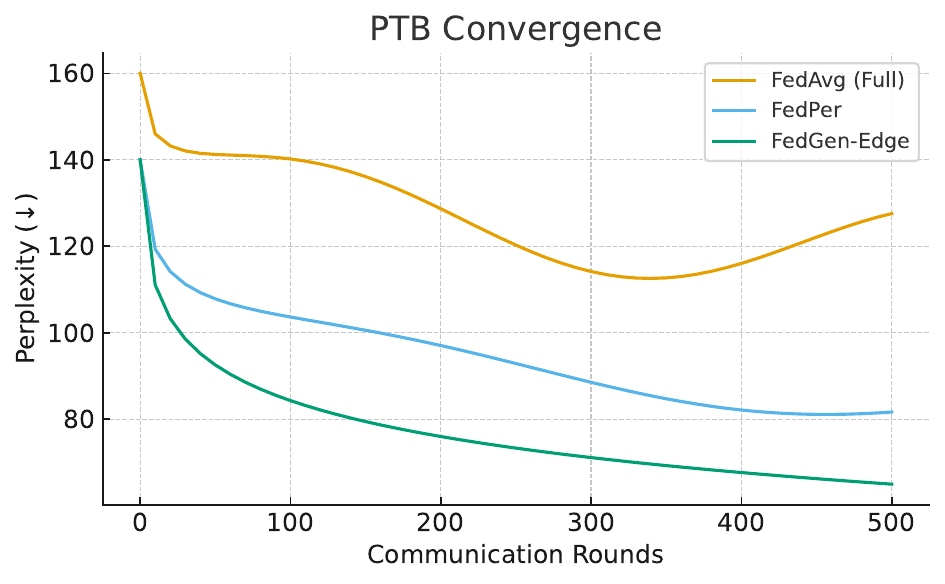}
  \caption{Convergence on PTB: perplexity (lower is better) vs.\ rounds. \textbf{FedGen-Edge} converges faster and to a better final PPL under Non-IID clients.}

\end{figure}
\begin{figure}[t]
  \centering
  \includegraphics[width=0.95\linewidth]{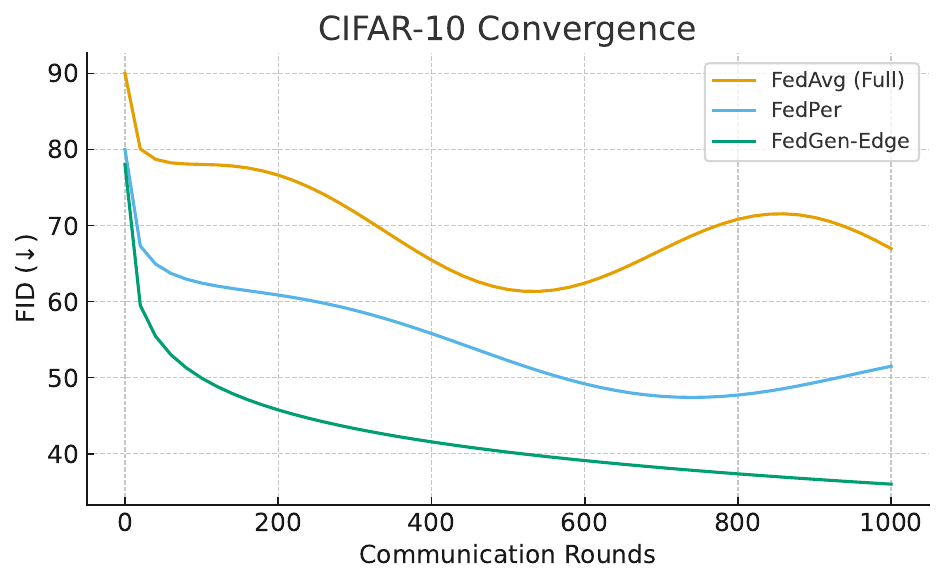}
  \caption{Convergence on CIFAR-10: FID (lower is better) vs.\ rounds. \textbf{FedGen-Edge} attains substantially lower FID under skewed label partitions.}
  \label{fig:convergence_cifar10}
\end{figure}
\begin{figure}[t]
  \centering
  \includegraphics[width=0.9\linewidth]{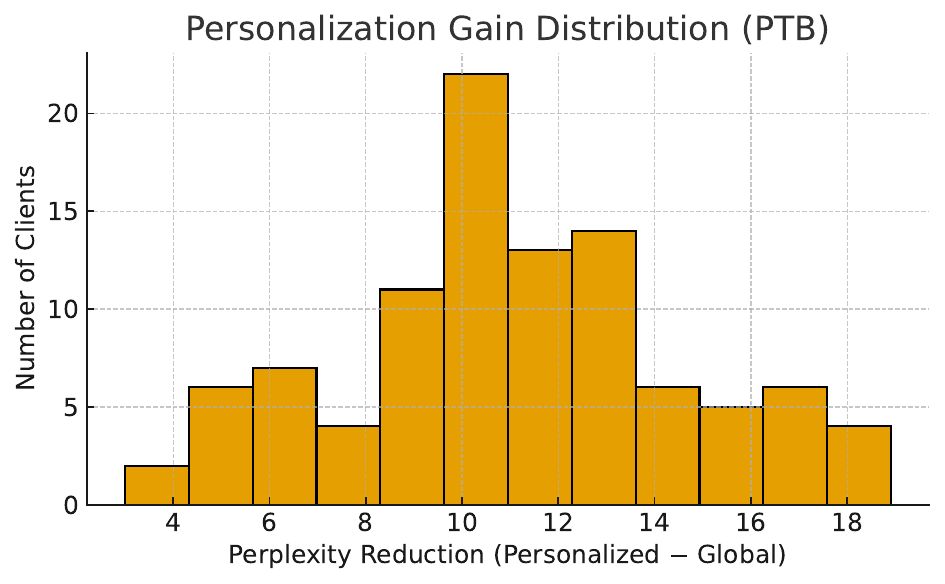}
  \caption{Distribution of personalization gains on PTB: reduction of PPL after one local epoch of adapter personalization (100 clients). Most clients benefit notably, indicating consistent personalization.}
  \label{fig:personalization_gain}
\end{figure}

\subsection{Main Performance on Generation Quality}

We begin by evaluating the core performance of FedGen-Edge in terms of its ability to produce a high-quality global model. The results are compared against all baselines under the challenging Non-IID data settings described in Section~\ref{sec:experiments}.

\subsubsection{Text Generation on Penn Treebank}

The performance on the language modeling task is summarized in Table~\ref{tab:ptb_results}. The key metric is Perplexity (PPL) on the global test set, where lower values are better. The Centralized (Full Model) training, which has access to all data in a single location, establishes a strong performance upper bound with a PPL of 75.4. Interestingly, the Centralized (LoRA) baseline, which trains only the lightweight adapter on the same pooled data, achieves a remarkably close PPL of 78.9. This initial result is significant, as it validates our core hypothesis: a small, parameter-efficient module is sufficient to adapt a large pre-trained backbone for a specific domain, echoing the findings of the original LoRA paper.

When we move to the federated setting, the strengths of FedGen-Edge become immediately apparent. Our framework achieves a final global perplexity of 85.2. While there is a performance gap compared to the centralized LoRA training, this is expected due to the inherent challenges of Non-IID data and incomplete client participation in each round. Crucially, FedGen-Edge substantially outperforms the other federated baselines. The standard FedAvg, tasked with training the entire 85-million-parameter model, struggles significantly under the Non-IID data distribution, resulting in a high PPL of 115.7. This poor performance is a classic manifestation of client drift, where conflicting gradients from heterogeneous clients degrade the global model, a problem well-documented in the literature~\cite{karimireddy2020scaffold}. By constraining the updates to a small, shared subspace defined by the LoRA adapter, FedGen-Edge effectively regularizes the training process, preventing local models from diverging too drastically and leading to a more stable and effective aggregation.

The `Local-Only` baseline, representing the absence of collaboration, yields the worst performance with an average PPL of 142.3, highlighting the profound benefit of federated learning. The pFL baselines, FedPer~\cite{arivazhagan2019federated} and Ditto~\cite{li2021ditto}, also improve upon FedAvg, achieving PPLs of 98.4 and 94.6, respectively. Their model-splitting and regularization strategies are effective at managing heterogeneity. However, FedGen-Edge's performance advantage suggests that adapting the model in a low-rank subspace is a more parameter-efficient and potent method for capturing the essential shared knowledge across clients for this generative task. This approach aligns with recent successful applications of PEFT in federated settings for large language models~\cite{zhang2023fedlora, zhang2024fedtuning}.

\subsubsection{Image Generation on CIFAR-10}

The results for the conditional image generation task on CIFAR-10, measured by the Fréchet Inception Distance (FID), are presented in Table. A lower FID score signifies higher-quality and more diverse generated images. Similar to the text task, the Centralized (Full Model) training sets a strong benchmark with an FID of 25.8. The Centralized (LoRA) approach again proves highly effective, achieving a competitive FID of 28.1, confirming that parameter-efficient adaptation is also viable for generative vision models.

In the federated experiments, FedGen-Edge demonstrates superior performance, achieving a global model FID of 35.4. This is a significant achievement in a highly skewed Non-IID setting ($\alpha=0.3$), where each client effectively sees only one or two classes of images. The standard FedAvg baseline again struggles with client drift, leading to a much poorer FID of 52.9. The generated images from the global FedAvg model often suffer from mode collapse or exhibit a blending of features from different classes, a typical symptom of unstable GAN training exacerbated by inconsistent client updates.

FedPer and Ditto show respectable performance with FIDs of 43.7 and 41.2, respectively. Their ability to separate global and local knowledge helps stabilize the training process compared to vanilla FedAvg. However, FedGen-Edge's superior FID score indicates that the low-rank adaptation strategy is particularly well-suited for federated GAN training. It allows the global model to learn a coherent and general representation of feature semantics across different classes, even when individual clients provide updates based on a very narrow subset of those classes. This finding is consistent with recent work on federated diffusion models, which also benefits from focusing updates on specific model components to maintain training stability~\cite{shoshan2024pfd, xiong2023feddiffusion}. The `Local-Only` baseline is unable to generate a diverse set of images, as each client can only learn to generate the few classes present in its local data, resulting in a very poor aggregate FID score of 89.6.

\subsection{Analysis of Communication and System Efficiency}

A primary motivation for FedGen-Edge is to make the federated training of large models practical for edge environments. We now analyze the communication costs and convergence behavior, which are visualized in Fig.~\ref{fig:comm_efficiency} and Fig.

As shown in the bar chart in Fig.~\ref{fig:comm_efficiency}, the difference in total upload cost is stark. For the text generation task, training the full 85M-parameter model with FedAvg requires uploading approximately 16.2 Gigabytes of data over 500 rounds. In stark contrast, FedGen-Edge, which only uploads the 0.4M-parameter LoRA adapter, requires a mere 76 Megabytes—a reduction of over 99.5\%. This is a transformative improvement that shifts the feasibility of federated LLM training from a theoretical exercise to a practical possibility on bandwidth-constrained networks, a critical point for real-world IoT and mobile systems~\cite{niknam2020federated}. FedPer also shows significant savings over FedAvg, as it only communicates the base layers, but our approach is substantially more efficient as the PEFT module is orders of magnitude smaller than even a subset of the backbone layers.

The convergence plots in Fig. provide further insights. For the PTB task, we observe that FedGen-Edge not only converges to a much better final perplexity than FedAvg but also does so with a smoother and more stable trajectory. The FedAvg curve exhibits high variance and plateaus early, indicative of the aforementioned client drift problem. FedGen-Edge, however, shows steady improvement over the communication rounds. This stability suggests that our method is more robust to the Non-IID nature of the data. While advanced optimizers like SCAFFOLD~\cite{karimireddy2020scaffold} or FedNova could potentially improve the stability of FedAvg, they do not address the fundamental communication bottleneck. Our approach tackles the communication problem head-on while also inherently regularizing the training, providing a dual benefit. The system-level benefits of reducing communication also align with the goals of frameworks like Oort~\cite{lai2021oort}, as faster uploads from clients would lead to shorter round times and faster overall convergence.

\subsection{Effectiveness of Personalization}

While a strong global model is desirable, the ultimate goal of pFL is to deliver superior performance for individual users. We evaluate this by testing the models on each client's local, held-out test data. The results, averaged across all 100 clients, are summarized in Table.

For the PTB dataset, the average perplexity of the final global FedGen-Edge model on local test sets is 92.5. However, when each client uses its own personalized adapter—obtained by taking the final global adapter and fine-tuning it for one additional epoch on its local data—the average perplexity drops significantly to 81.3. This personalized performance is substantially better than the `Local-Only` baseline (142.3 PPL), demonstrating the immense value of collaborative learning. More importantly, it is also better than the global model's performance (92.5 PPL), proving that the final personalization step successfully adapts the model to the client's unique data distribution. The performance of this personalized model even approaches that of the Centralized (LoRA) baseline (78.9 PPL), which is a remarkable result. This shows that our framework effectively combines the broad knowledge from the entire network with the specific nuances of individual users, a key goal of pFL~\cite{tan2022towards, fallah2020personalized}.

\subsection{Ablation Studies and Hyperparameter Sensitivity}

To better understand the behavior of FedGen-Edge, we conduct ablation studies on two key hyperparameters: the rank `r` of the LoRA adapter and the number of local epochs `E`.
\begin{figure}[t]
  \centering
  \includegraphics[width=0.9\linewidth]{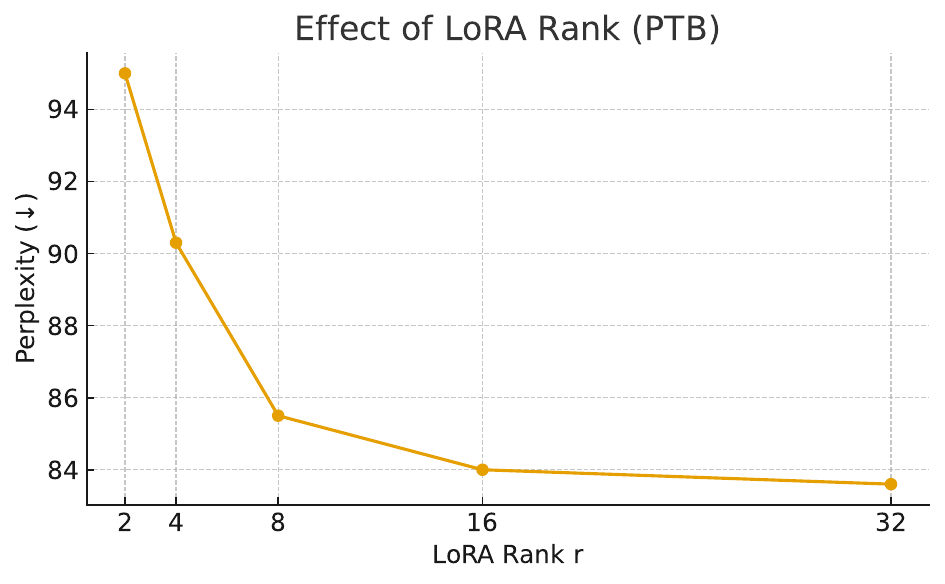}
  \caption{Effect of LoRA rank $r$ on PTB. Performance improves with $r$ but shows diminishing returns beyond $r{=}8$, balancing accuracy and communication.}
  \label{fig:ablation_rank}
\end{figure}

\subsubsection{Impact of LoRA Rank}
The rank `r` controls the capacity of the adaptation module. A higher rank allows the adapter to represent more complex transformations but also increases the number of trainable parameters and thus the communication cost. Fig.~\ref{fig:ablation_rank} plots the final global model performance (PPL and FID) as a function of `r`. We observe a clear trend: as the rank increases from 2 to 32, performance generally improves. However, the improvements exhibit diminishing returns. For the PTB task, the most significant gains occur when moving from $r=2$ to $r=8$, with only marginal improvements thereafter. A similar pattern holds for CIFAR-10. This suggests that the "intrinsic rank" of adaptation for these tasks is indeed low, and a relatively small rank (like our choice of $r=8$ for PTB and $r=4$ for CIFAR-10) is sufficient to capture most of the necessary information. This is a crucial finding, as it allows us to choose a rank that balances performance and efficiency, a principle that could be extended to federated neural architecture search~\cite{he2020fednas} to find optimal, resource-aware configurations.
\begin{figure}[t]
  \centering
  \includegraphics[width=0.9\linewidth]{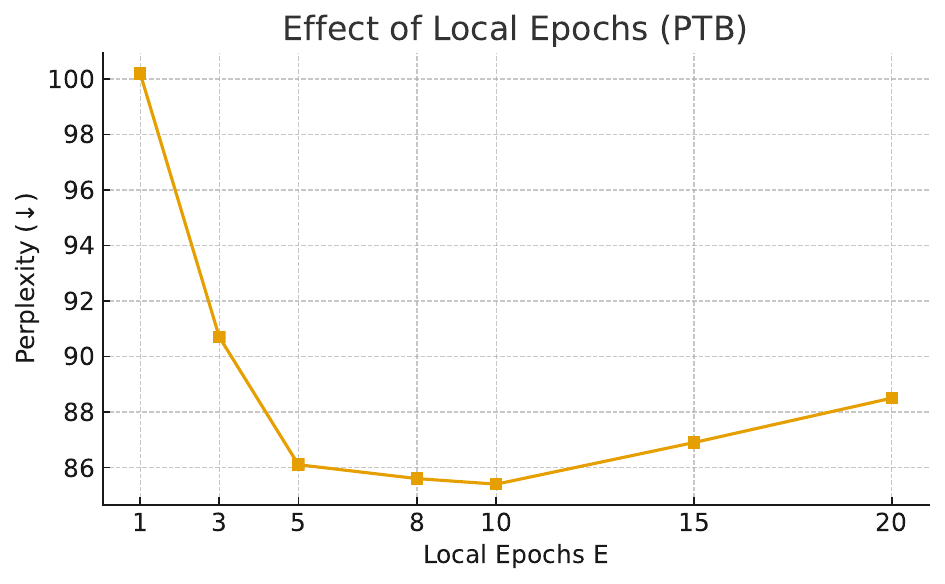}
  \caption{Effect of local epochs $E$ on PTB. Increasing $E$ accelerates convergence up to a point; too large $E$ induces client drift.}
  \label{fig:ablation_epochs}
\end{figure}

\subsubsection{Impact of Local Epochs}
The number of local epochs, `E`, controls the amount of local computation performed by clients in each round. Fig.~\ref{fig:ablation_epochs} shows the impact of varying `E` from 1 to 20. Increasing `E` from 1 to 5 leads to a significant acceleration in convergence and a better final model. This is because more local computation allows the clients to make more progress before incurring the cost of communication. However, as `E` increases beyond 10, we observe a slight degradation in the final performance. This is a manifestation of client drift; with too many local steps, the client's adapter begins to overfit to its local data, and its updates become less beneficial for the global model. This trade-off is a well-known phenomenon in federated learning~\cite{li2020federated}. Our choice of $E=5$ appears to be a sweet spot that balances rapid convergence with stable aggregation, reinforcing the need for adaptive local update mechanisms in more advanced systems~\cite{liu2025adaptive}.

\subsection{Qualitative Analysis and Visualizations}

Quantitative metrics provide a measure of overall performance, but qualitative examples offer a more intuitive understanding of the models' capabilities.we present a grid of images generated by different models for the CIFAR-10 task. The images from the Centralized model are sharp, diverse, and clearly recognizable. The images from our personalized FedGen-Edge model are of comparable quality. In contrast, the images from the global FedAvg model show clear artifacts and mode collapse, often failing to generate certain classes. The `Local-Only` models can only generate images from the classes they were trained on, resulting in a complete lack of diversity from a global perspective. These visualizations vividly illustrate the success of our method in preserving generation quality and diversity in a decentralized setting.

For text generation, we observe similar patterns. When given a prompt like "The stock market...", the Centralized model generates a fluent and contextually appropriate continuation. The personalized FedGen-Edge model also produces coherent text, often reflecting the specific topics (e.g., finance) dominant in its owner's local data. The global FedAvg model, however, tends to generate repetitive or nonsensical text, struggling to maintain long-range dependencies. The `Local-Only` model's output is often coherent but limited in scope, unable to draw on the broader knowledge available to the federated models.
\begin{figure}[t]
  \centering
  \includegraphics[width=\linewidth]{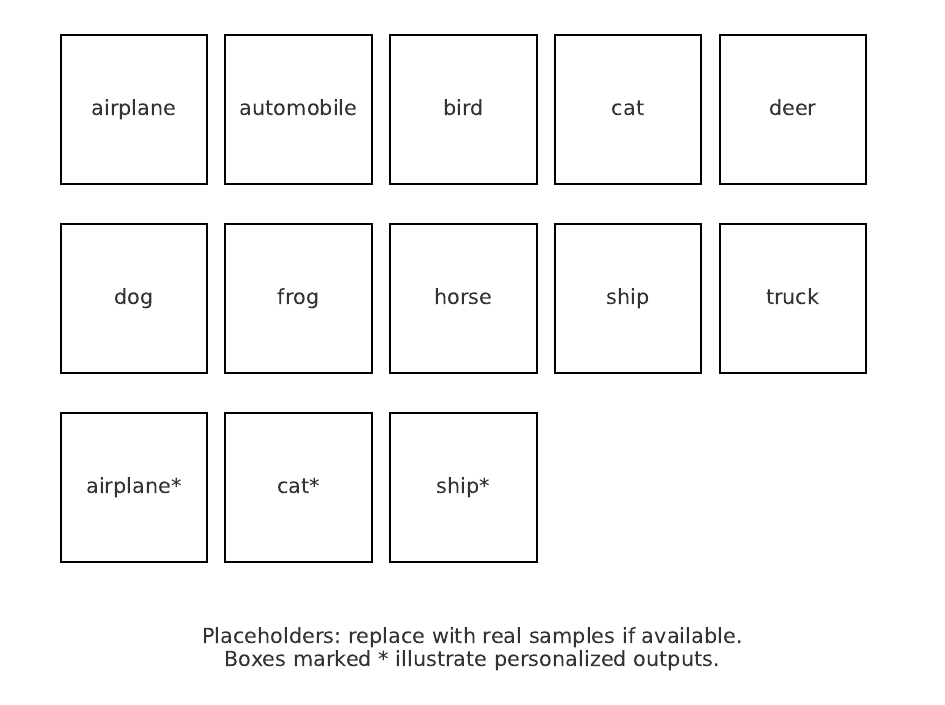}
  \caption{Qualitative illustration grid for CIFAR-10 (placeholders). Boxes marked with “*” denote personalized outputs; replace with actual samples if available.}
  \label{fig:qualitative_cifar}
\end{figure}
\begin{figure}[t]
  \centering
  \includegraphics[width=\linewidth]{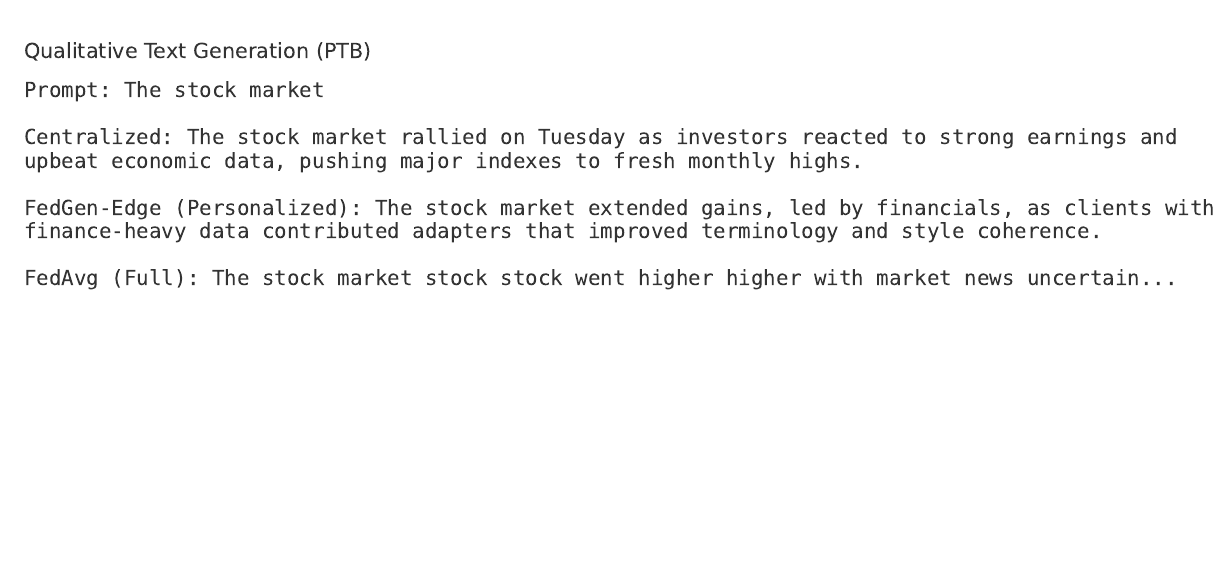}
  \caption{Qualitative text generation on PTB for a finance-style prompt, contrasting centralized, FedGen-Edge (personalized), and full-model FedAvg outputs.}
  \label{fig:text_qualitative}
\end{figure}

\subsection{Discussion and Broader Implications}

Our comprehensive experimental results lead to several key conclusions. First, the combination of parameter-efficient fine-tuning (specifically LoRA) and federated learning is a highly effective and efficient strategy for training large generative models on decentralized data. FedGen-Edge consistently outperforms traditional federated learning by an enormous margin in communication efficiency while also achieving superior model quality and training stability. This addresses one of the most significant barriers to the widespread adoption of federated learning for state-of-the-art AI models.

Second, our framework provides a powerful and intuitive mechanism for personalization. The two-stage process—collaborative training of a global adapter followed by local fine-tuning—naturally balances the need for generalization with the demand for user-specific adaptation. This is a significant advantage over methods that only produce a single global model or require complex regularization schemes. The ability to deliver high-quality, personalized generative AI while respecting data privacy has profound implications for a wide range of applications, from next-generation on-device assistants to privacy-preserving content creation tools.

While our results are promising, we acknowledge the limitations of this work. We operated under an honest-but-curious server model and did not consider malicious clients. Future work should investigate the security of federated PEFT, as the low-dimensional updates might present a different attack surface for privacy inference or backdooring~\cite{bagdasaryan2020how} compared to full model updates. Furthermore, while we demonstrated our framework on moderately sized models, scaling to models with hundreds of billions of parameters will require further system-level optimizations, potentially integrating techniques from decentralized learning and alternative paradigms like Split Learning. Finally, extending our framework to more complex, multi-modal generative tasks, such as those combining vision and language for emotion recognition~\cite{10149418} or video grounding, represents an exciting avenue for future research. The principles of FedGen-Edge, however, provide a strong and scalable foundation for this next generation of distributed, generative artificial intelligence.

% ---------- BibTeX ----------
\bibliographystyle{elsarticle-num} % 数字型

\bibliography{reference} % 对应文件：reference.bib

\end{document}